\documentclass[letterpaper, 10pt,conference]{ieeeconf}
\IEEEoverridecommandlockouts
\usepackage{times,mathptmx,latexsym}
\usepackage[T1]{fontenc}
\DeclareFontShape{T1}{ptm}{m}{scit}{<->ssub * ptm/m/it}{}
\usepackage[utf8]{inputenc}
\DeclareUnicodeCharacter{2010}{-}
\usepackage{microtype,inconsolata,graphicx,booktabs,multirow}
\microtypesetup{protrusion=false}
\usepackage{amsmath,amssymb,mathtools}
\usepackage{pgfplots}
\pgfplotsset{compat=1.18}
\usepackage{cite,xurl}
\makeatletter
\let\NAT@parse\undefined
\makeatother
\usepackage[hidelinks]{hyperref}
\hypersetup{pdftitle={Instruction-Conditioned Motion Control for Robotic Arms: Language-to-Code Decomposition and Reinforcement Learning}}
\usepackage[capitalize,noabbrev]{cleveref}
\crefname{figure}{Fig.}{Figs.}
\usepackage{fvextra}
\title{\LARGE\bfseries Instruction-Conditioned Motion Control for Robotic Arms:\\
Language-to-Code Decomposition and Reinforcement Learning}
\author{Fan Huang$^{1,*}$\thanks{\textsuperscript{*}Corresponding author: \href{mailto:huangfan@acm.org}{\texttt{huangfan@acm.org}}}, Dezhong Tong$^{2}$, Changyu Chen$^{3}$, Haewoon Kwak$^{1}$, and Jisun An$^{1}$\\
\small $^{1}$Indiana University Bloomington\quad
$^{2}$University of Michigan\quad $^{3}$Singapore Management University}
\hypersetup{pdfauthor={Fan Huang, Dezhong Tong, Changyu Chen, Haewoon Kwak, Jisun An}}

\IEEEaftertitletext{\vspace{-\baselineskip}}
\begin{document}
\bstctlcite{IEEEexample:BSTcontrol}
\maketitle
\thispagestyle{plain}
\pagestyle{plain}
\raggedbottom

\begin{abstract}
Instruction-Conditioned Motion Control (ICMC) separates parameter specification, instruction generation and code synthesis for language-driven robot control, making execution outcomes traceable to instructions and programs. We compare one-shot generation with ICMC under o1 and o4-mini on cube pick-and-place and letter-reshaping tasks for a UR5e arm in MuJoCo. We then train an open-weight instruction policy by GRPO through a frozen code generator on cube placement onto two- and five-cube letter layouts, averaging the rewards of five programs per instruction. The logged positive-reward rate rose from 31.6\% to 35.3\% (253/800 to 282/800) between the first and last five training steps. A log audit identifies positive binary labels associated with geometric-failure messages: at least 19 in that run, 18 in the dense-reward run and 5 in the run with one program per instruction. In one case, all five rewards were 1.0 while the log reports a 0.9515\,m miss against a 0.05\,m tolerance; the verdict came from the program's own output stream. We distinguish logged reward rates from independently verified task success and report descriptive results for each configuration's single training trajectory. We introduce an independent evaluation protocol that checks task completion against fixed specifications, separating robot control from success assessment to support reproducible evaluation.
\end{abstract}

\section{Introduction}
Language models can generate programs that invoke robot primitives, so that a manipulation request can be expressed through a familiar code interface \cite{codeaspolicies,progprompt,chatgptrobotics}. Prior work grounds actions in robot capabilities and current state \cite{saycan,progprompt} and uses feedback to recover from failures \cite{innermonologue,chatgptrobotics}, while rearrangement evaluation compares the achieved scene with a specified goal \cite{batra2020rearrangement}. These considerations go beyond syntactic correctness and motivate inspecting how a task becomes an executable program.

Two issues motivate this study. First, our one-shot baseline produces a complete program without a separately generated instruction artifact. Prior work combines code with execution-grounded reasoning \cite{chainofcode}; we investigate a trainable instruction stage for tracing downstream failures. Second, one stochastic code sample can confound instruction quality with sampling variation, motivating our reward-averaging analysis. Separately, imperfect reward signals can favor unintended behavior or diverge from intended performance \cite{amodei2016safety,gao2023overoptimization}. Instruction-Conditioned Motion Control (ICMC) separates parameter specification, instruction generation and code synthesis over a fixed robot API for a UR5e arm in MuJoCo \cite{mujoco}. The intermediate action is the entire generated instruction text, which in practice contains both prose and a reference-code block (median code-line fraction 79.3\% across the 32 instructions of the final step of the main binary-reward training run), so it is a language-to-code interface rather than a short symbolic plan.

We ask three questions. RQ1: how do one-shot code generation and ICMC compare under o1 and o4-mini on cube pick-and-place and letter reshaping, and how do same-backbone Chain-of-Code and iterative-correction implementations compare on reshaping? RQ2: what happens when an open-weight instruction policy is trained by GRPO \cite{grpo} through a frozen code generator on cube placement onto letter-shaped layouts, with rewards parsed from program execution? RQ3: what do the archived comparisons of reward sampling (M=5 versus M=1), learning rate and update epochs establish, and which effects remain confounded?

This paper makes four contributions. First, a language-to-code decomposition over a fixed UR5e API that trains an inspectable instruction policy through a frozen code generator; retained records link instructions to syntax errors, undefined names and discrepancies between reward labels and logged geometry. Second, an audit of reward consistency: the main run's logged positive-reward rate rose from 31.6\% to 35.3\% between the first and last five steps, while inspection identifies at least 19 positive-reward logs reporting geometric failure in that run, 18 in the dense-reward run and 5 with one program per instruction (M=1). The audit traces success labels to the generated program's output stream. Third, a characterization of downstream reward variance: an estimator argument for averaging M program rewards per instruction, supported by the observation that 302 of the 640 instructions in the main run received mixed rewards across their five programs. Fourth, an independent evaluation protocol for checking task completion against fixed specifications, with simulator state and scoring controlled by a separate software supervisor. Its CPU implementation passed ten negative controls; a hand-authored reference controller completed all 40 predeclared scenes under this protocol.

The study characterizes simulated execution through one archived training trajectory per configuration. Evaluation of model-generated programs under the independent protocol, replication across training seeds, and matched tests of learning-rate and epoch effects remain future work.

\begin{figure*}[t]
\centering
\includegraphics[width=\textwidth]{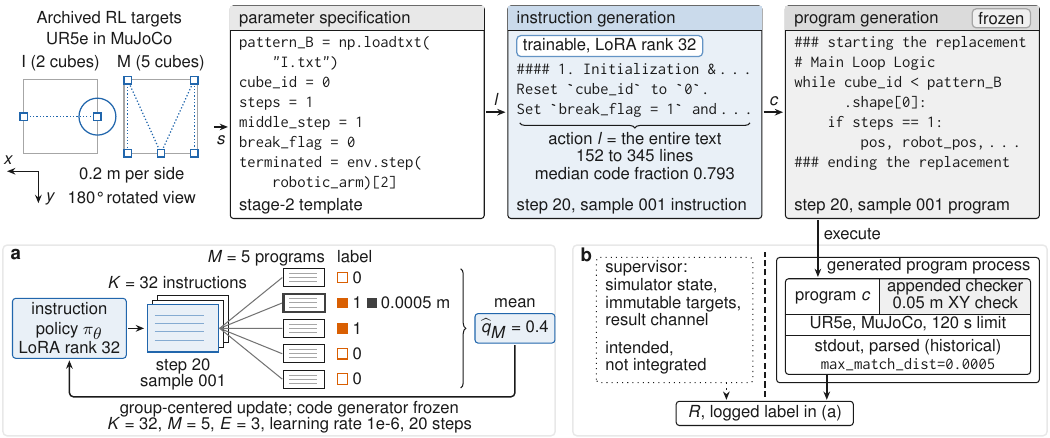}
\caption{Instruction-conditioned motion control. Top: archived I (2 cubes) and M (5 cubes) targets in world XY rotated $180^\circ$, each in a 0.2\,m square; circle radius 0.05\,m. Dotted connections are visual guides. The stages show template and step 20, sample 001 excerpts. (a) Five programs yield logged labels 0, 1, 1, 0, 0 and mean 0.4; only the instruction policy is updated. (b) The appended checker shares the generated program's process and stdout. Archived training used the shared-process checker; independent task verification is specified for future model evaluations. TikZ drafted with Anthropic Claude and revised with OpenAI Codex; initial prompt excerpt: ``Build a stronger Figure 1 as a standalone TikZ document.''}
\label{fig:pipeline}
\end{figure*}

\section{Related Work and Scope}
Code-as-Policies \cite{codeaspolicies} uses hierarchical code generation, and ProgPrompt \cite{progprompt} generates program-like task plans with execution-time assertions; Chain-of-Code \cite{chainofcode} interleaves reasoning with execution, and ChatGPT for Robotics \cite{chatgptrobotics} studies robot-specific prompting and feedback. ICMC inserts an instruction stage between the task and the program and trains that stage with execution reward while the coder stays fixed, using group-based policy optimization \cite{grpo} with LoRA \cite{lora}. Vision-language-action policies and embodied language models, including RT-1, RT-2, PaLM-E and OpenVLA \cite{rt1,rt2,palme,openvla}, and affordance-grounded or closed-loop language systems \cite{saycan,innermonologue}, use observation and action interfaces distinct from our text-only instruction policy; we discuss them as related approaches.

Transporter Networks \cite{zeng2020transporternets} learn pick/place poses from RGB-D demonstrations in Ravens. CLIPort \cite{cliport} adds language-conditioned semantic features to spatial imitation learning, using task-dependent partial-credit scores. Both use manipulation-policy interfaces distinct from Python-program generation. Rearrangement frameworks \cite{batra2020rearrangement} distinguish goal specification, sensing, actuation and evaluation. Our text-only RL inputs, generated-code action interface, letter tasks and intended all-object criterion differ on each of those axes.

\section{Method}
\subsection{Instruction-conditioned control}
Let $s=(\tau,o)$ contain a task description and available scene information. An instruction policy generates $l\sim\pi_\theta(\cdot\mid s)$; a frozen code policy generates $c\sim\pi_{\rm code}(\cdot\mid s,l)$. A program invokes a fixed robot interface inside the simulator. The conceptual objective is
\begin{equation}
J(\theta)=\mathbb E_{s,l,c}[R(c,s)],
\end{equation}
where a task-specific evaluator determines $R$. In the archived runs $R$ was parsed from program output; \cref{sec:boundary} explains how these labels are interpreted.

Stage 1 specifies task parameters, robot configuration, object identifiers, target coordinates and operational context. Stage 2 generates instructions for perception, fetching and placing, including state transitions. Stage 3 receives those instructions with the API and reference code and generates a complete program. The whole stage-2 token sequence is the trainable action; stage 3 stays frozen. Exposing the intermediate artifact helps identify whether a downstream program ignores an instruction, changes a function contract, or terminates before reaching the requested scene. \Cref{fig:pipeline} summarizes the three stages.

The simulator interface includes cube-pose lookup, gripper opening/closing and trajectory execution. A trajectory must follow the implementation's $(N, 8)$ action convention. Reference state variables include \texttt{steps}, \texttt{middle\_step}, \texttt{cube\_id} and \texttt{break\_flag}. These implementation details constrain the form of generated programs.

\subsection{Reward sampling and optimization}
For a fixed task $s$ and binary evaluator, let $q(l)$ be the probability that a fresh downstream sample given instruction $l$ receives label 1. With conditionally independent samples from that same distribution,
\begin{align}
\widehat q_M(l)&=\frac1M\sum_{m=1}^{M}R(c_m,s),\\
\mathbb E[\widehat q_M\mid l]&=q(l),\qquad
\operatorname{Var}(\widehat q_M\mid l)=\frac{q(l)(1-q(l))}{M}.
\end{align}
This estimates the chosen evaluator's label probability, which represents task success only if the evaluator is valid. The variance reduction is conditional on the instruction and costs M times the nominal downstream sampling budget. The archive shows that the variance it targets is present: of the 640 instructions sampled in the main binary run, 269 received five zero rewards, 69 five positive rewards and 302 mixed rewards, so for 302 instructions (47.2\%) a single program would have assigned a reward different from the five-program mean. These counts characterize logged-label variation; attributing training-curve changes to specific noise sources requires controlled evaluation.

For K instruction samples, a centered reward coefficient is $A_k=\widehat r_k-K^{-1}\sum_j\widehat r_j$. The score-function component has direction
\begin{equation}
g(\theta)=\mathbb E\!\left[\sum_k A_k\nabla_\theta\log\pi_\theta(l_k\mid s)\right].
\end{equation}
This is the update direction before normalization, clipping and regularization; a group whose K rewards are equal contributes no centered signal. The archive does not identify the source version governing clipping, the KL coefficient or advantage normalization, so we report the update at this level.

The archived runs combine M=5 averaging with learning rate $10^{-6}$ and E=3 update passes on each sampled batch. The learning rate and the number of passes are empirical choices. Binary and dense variants were trained separately. The dense reward is reported as
\begin{equation}
r=R-0.3V_{\rm bug}+0.25r_{\rm goal}+0.25r_{\rm comp},
\end{equation}
where $V_{\rm bug}$ averages signature, scope and literal-index heuristics, $r_{\rm goal}$ reflects distance progress, and $r_{\rm comp}$ grades completion. The detector is heuristic: precision and recall remain to be measured, and false positives can penalize valid programs. Reward-hacking risks in the shaped terms motivate independent evaluation \cite{amodei2016safety,gao2023overoptimization}.

\section{Tasks, Configurations and Evaluation}
The following terms are used throughout. An archived run is one of the runs whose records this paper analyzes; its scorer parsed program output, and a logged rate is the fraction of observed binary labels equal to 1, including the binary component of the dense reward. Validated success is reserved for geometric checks outside the generated program; the archived runs provide labels parsed from program output. The denoised reward is the mean of M=5 program rewards for one instruction. The prompt-fix baseline is the instruction policy without LoRA training, given the same API addendum, on the M task, with an instruction-condensing call before code generation that the training runs do not have. I and M name the archived cube layouts: training uses the two- and five-cube targets in \cref{fig:pipeline}; the benchmark suite uses distinct nine- and thirteen-cube versions. The single-epoch runner is a later re-implementation used only in the learning-rate and epoch comparison. The preliminary sweep is an earlier set of seven trained settings and two no-RL controls whose configurations vary in group size, horizon, reward and instruction format.

\subsection{Environment, tasks and configurations}
\begin{figure}[t]
\vspace*{4pt}
\centering
\includegraphics[width=\columnwidth]{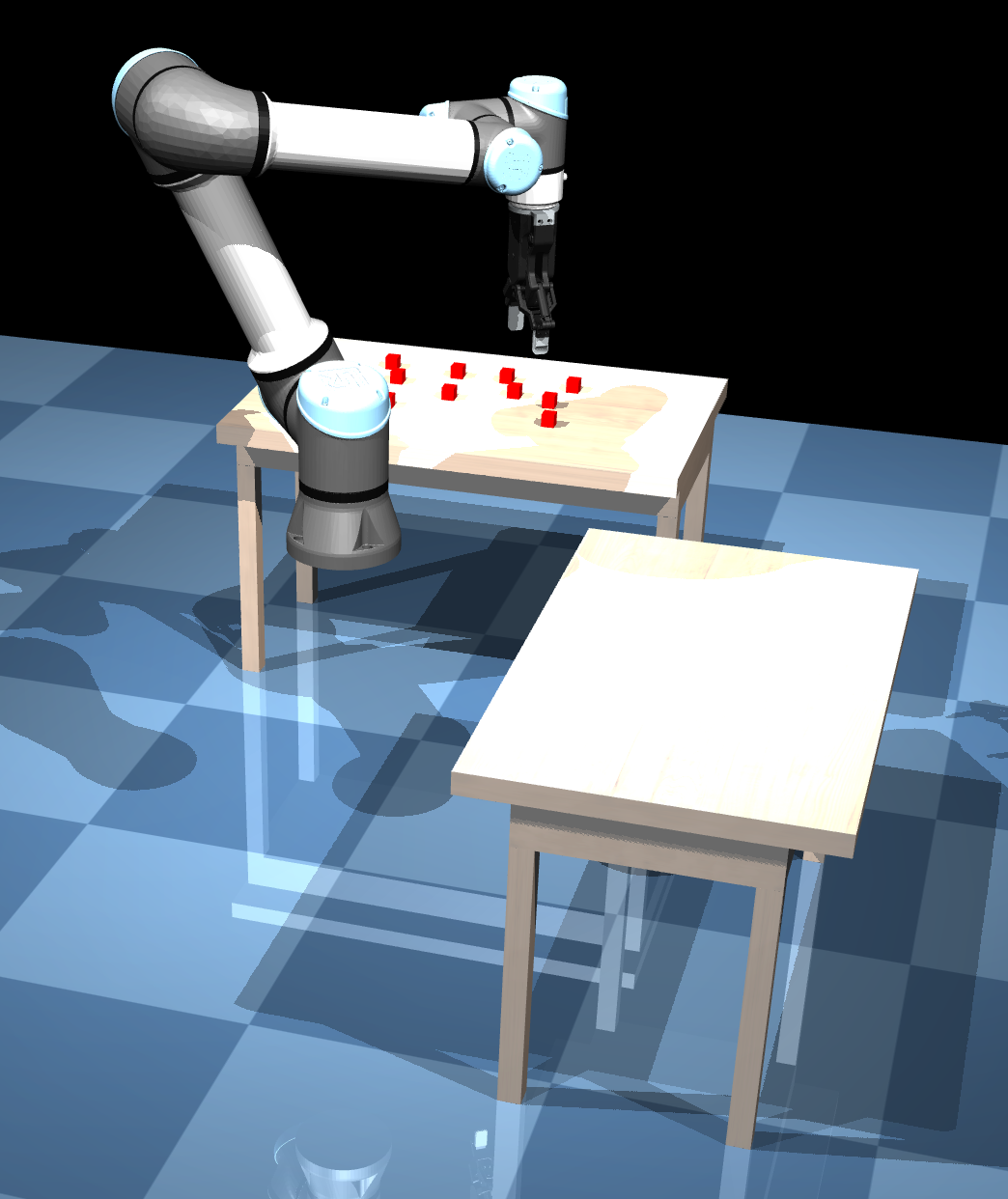}
\caption{MuJoCo Simulation Environment. A UR5e arm with parallel gripper manipulates red cubes on the rear table and transfers them to the disposal area on the front table.}
\label{fig:mujoco-env}
\end{figure}

The study uses MuJoCo \cite{mujoco} through a Gymnasium interface \cite{gymnasium}, with a UR5e arm and parallel gripper (\cref{fig:mujoco-env}). RQ1 includes image-conditioned prompting; RQ2/RQ3 use textual task/scene descriptions.

Pick-and-place moves selected cubes to target positions or a disposal area. Shaping generates letter coordinates; its workspace-bounds metric is distinct from correct letter geometry and from robot execution. Reshaping introduces removals, additions and assignments between source and target coordinates. A defined suite contains 12 targets and therefore 66 unordered pairs, or 132 directed transformations.

A task-distance score $D(A,B)=|A\triangle B|+6\mathbf1_{\rm encl}+3\mathbf1_{\rm sym}$ bands the 66 pairs by coordinate-set difference and changes in archived enclosure and symmetry labels; on the 16 planning pairs its Spearman correlation with an assignment-derived operation proxy is 0.919, and the coordinate-set term alone gives 0.975. The score is symmetric and is not a measure of execution difficulty.

RQ1 reports o1 and o4-mini prompting configurations; the o1 execution cells contain five trials and o4-mini twenty. RQ2/RQ3 report a Qwen3.5-35B-A3B instruction policy with rank-32 LoRA ($\alpha=64$) and a frozen Qwen3-Coder-Next service through OpenRouter. Central settings are K=32, M=5, E=3, 20 steps, instruction/code temperatures 0.7/0.1, a 4096-token generation limit and 120-second execution timeout.

Offline recovery establishes 60 saved o4-mini coordinate outputs (12 shapes, five each), all within strict workspace bounds. A separate archive has 80 plans over 16 directed pairs; 76 satisfy an explicit coverage/uniqueness rubric, and 38 also satisfy maximal reuse. The second condition is a stricter preference layered on the first, and both count plans rather than executions. In the small five-to-two-node example, all five o1 and five o4-mini plans are coverage-feasible; adding maximal reuse gives 4/5 and 5/5. In this example the Hungarian solver and the listed model assignments have the same padded objective (0.653); the smaller real-to-real subtotal quoted for the model plans measures a different quantity.

\subsection{Evaluation protocol and interpretation of logged rewards}\label{sec:boundary}
A geometric criterion requires an explicit correspondence between final objects and immutable targets, a coordinate frame, tolerance boundary, object-count rule, and removal/addition protocol. The archived descriptions state a 0.1\,m tolerance for RQ1 and a 0.05\,m XY tolerance for RQ2/RQ3. The tolerances bound placement error; comparison with cube size requires version-specific cube dimensions that are not established by the archive.

\begin{figure}[t]
\vspace*{3pt}
\centering
\includegraphics[width=\columnwidth]{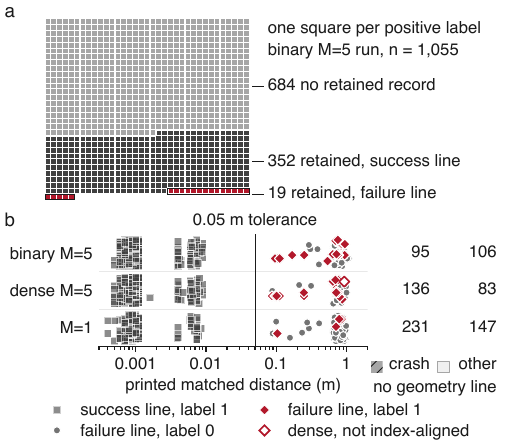}
\caption{Evidence behind a positive logged label. (a) The 1,055 positive labels of the binary M=5 run: 684 with no retained record, 352 with the checker success line, 19 with a geometric-failure message. (b) Largest printed matched distance per record against the 0.05\,m tolerance; failure logs may stop at the first violation. The open diamond (dense step 13, sample 026) has a compacted component list and is excluded from the index-aligned 18. Right: records without a geometry line. One program is retained per instruction (one of five for M=5; one for M=1), so retention differs across M. Success messages require independent verification. Each configuration has one archived run.}
\label{fig:label_vs_geometry}
\end{figure}

In the archived runs, success and geometric-error messages were parsed from the program's standard output, and process exit and timeout status marked execution failures. In the retained records the placement checker is appended by the runner to the generated program and runs inside the program's process, so its verdict shares an output stream with whatever the program prints; the archive does not establish which scoring-source version was used for each run. The retained-record scan finds at least 19 positive rewards whose retained log contains an explicit geometric-failure message in the main binary run, 18 in the dense run and 5 in the M=1 run (\cref{sec:examples}). Success-line records (352, 343, 163) print at most 0.009\,m, failure-line records (87, 78, 99) at least 0.0874\,m, and the 19, 18 and 5 positively labeled failure records 0.0874 to 0.9515\,m (\cref{fig:label_vs_geometry}). These are lower bounds: the driver kept only the first reward-maximizing program per instruction, so the 19 occur among the 371 instructions of the main run with at least one positive logged reward and the 5 among the 168 such instructions of the M=1 run, and the other programs behind each averaged reward were not saved. The single-epoch runner used in the learning-rate and epoch comparison, like the current source of the main runner, falls back to accepting any output that contains \texttt{SUCCESS} and no \texttt{FAIL}; whether the archived runs executed that source is not established. Removing these visible contradictions from the counts leaves the unsaved programs unresolved and does not change the policy trained on the original rewards.

The proposed independent protocol assigns simulator state, immutable task data and result reporting to a separate software supervisor. Generated programs issue control actions; the supervisor computes and reports task success. The protocol adopted for new evaluations inventories all ten cubes and predeclares placement, addition, disposal and reserve roles; primary success requires task placement, required disposal, reserve preservation and clean execution, with diagnostics reported separately, and reserves cannot substitute for task cubes. Every method receives the same specification.

The CPU implementation of this protocol passed ten negative controls and five predeclared 120\,s no-action holds under a 0.1\,mm drift rule (with a 1\,mm XYZ reserve tolerance for the pinned setup), and a hand-authored reference controller passed 40 of 40 predeclared scenes (20 I, 20 M) under it at 1\,cm placement tolerance, with no case replacement or controller adaptation. These results characterize the evaluation infrastructure. Evaluation of model-generated programs, target-host validation and re-scoring retained programs are the next steps. Such re-scoring covers one program per instruction; reconstructing averaged rewards would require the unsaved samples.

\section{Prompting Comparisons}
\begin{table}[t]
\vspace*{4pt}
\caption{Reported execution counts (successes/trials) for one-shot code generation and ICMC. The one-shot cells were recounted from the saved notebook status labels; the ICMC cells are the archived report values, of which only the o4-mini reshaping cell has a notebook recount (see the text at \cref{tab:reshaping}).}
\label{tab:prompting}
\centering\small
\begin{tabular}{@{}llcc@{}}
\toprule Task & Model & One-shot & ICMC\\\midrule
Pick/place & o1 &5/5&5/5\\
Pick/place & o4-mini&7/20&12/20\\
Reshaping & o1&1/5&3/5\\
Reshaping & o4-mini&5/20&3/20\\\bottomrule
\end{tabular}
\end{table}
\Cref{tab:prompting} compares one-shot code generation with ICMC on pick-and-place and reshaping, with five trials per o1 cell and twenty per o4-mini cell. The o1 pick-and-place cells tie at 5/5. With o4-mini, ICMC records 12/20 against 7/20 for one-shot on pick-and-place (Wilson 95\% intervals [38.7, 78.1]\% and [18.1, 56.7]\%; Fisher exact p = 0.205), and one-shot records 5/20 against 3/20 on reshaping. These results characterize their respective task protocols and model configurations; isolating reasoning effects across models requires matched conditions.

\begin{figure}[t]
\centering
\begin{tikzpicture}
\begin{axis}[ybar,bar width=6pt,width=0.95\columnwidth,height=4cm,
 ymin=0,ymax=110,ylabel={Reported proportion (\%)},
 symbolic x coords={CoC,Iter.,Base,Func.,Prompt,Both},xtick=data,
 tick label style={font=\footnotesize},label style={font=\footnotesize},
 legend style={font=\footnotesize,at={(0.5, 1.02)},anchor=south,legend columns=2},
 enlarge x limits=0.1]
\addplot coordinates{(CoC, 55)(Iter., 15)(Base, 75)(Func., 100)(Prompt, 70)(Both, 90)};
\addplot coordinates{(CoC, 20)(Iter., 5)(Base, 15)(Func., 20)(Prompt, 35)(Both, 50)};
\legend{Code-valid,Task-success}
\end{axis}
\end{tikzpicture}
\caption{Reported code validity and task success from \cref{tab:reshaping} for the six o4-mini reshaping configurations (n=20 each). Configurations use different task sets.}
\label{fig:reshaping}
\end{figure}
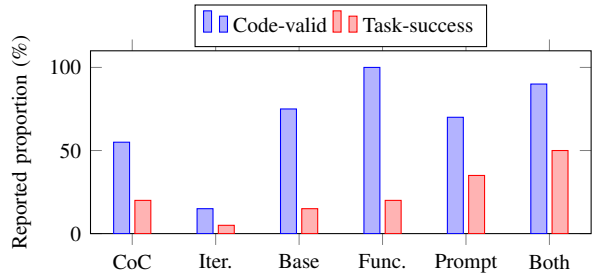

\begin{table}[t]
\vspace*{4pt}
\caption{Archived o4-mini reshaping outcomes (n=20 per row). The first two rows are Chain-of-Code and iterative-correction implementations on the same backbone; the ICMC rows differ in function set and prompt. Baseline and ICMC rows use different task sets and generation budgets; the comparison is descriptive.}
\label{tab:reshaping}
\centering\small
\begin{tabular}{@{}lrr@{}}
\toprule Configuration & Code-valid & Task-success\\\midrule
Chain-of-Code implementation&55\%&20\%\\
Iterative correction&15\%&5\%\\
Base ICMC&75\%&15\%\\
ICMC + amended functions&100\%&20\%\\
ICMC + constrained prompt&70\%&35\%\\
ICMC + both&90\%&50\%\\\bottomrule
\end{tabular}
\end{table}
\Cref{tab:reshaping} separates base ICMC from two engineering changes; the final row reports ICMC with both modifications. The iterative implementation allows an initial generation plus three retries and is scored on the final attempt; its any-attempt success is also 1/20. The 40 baseline runs all request the M-to-I transformation, whereas the ICMC notebooks mix four pairs from the Easy band of the task-distance score and include manual reruns, so the baseline rows and the ICMC rows do not share a task set (\cref{fig:reshaping}). Under a per-file recount of the notebooks, the notebook matching the base ICMC configuration yields 14/20 code-valid and 2/20 explicit success messages, and the final row yields 4/20 or 13/20 depending on whether the first or last saved manual execution is taken. The table keeps the archived report values; code validity in some rows counts timeouts as valid.

Wilson 95\% intervals for n=20 cells are wide: base ICMC at 3/20 has [5.2, 36.0]\%, the combined ICMC row at 10/20 has [29.9, 70.1]\% and Chain-of-Code at 4/20 has [8.1, 41.6]\%, so these three task-success intervals overlap pairwise. These intervals summarize individual proportions; testing differences between rows remains future work.

\section{Training Through a Frozen Code Generator}
\begin{figure}[t]
\vspace*{4pt}
\centering
\includegraphics[width=\columnwidth]{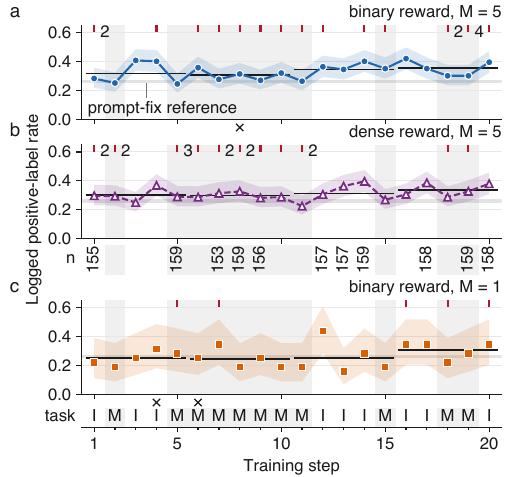}
\caption{Logged positive-label rates for binary M=5 (a), dense M=5 (b) and binary M=1 (c), one trajectory each. Ribbons: descriptive Wilson 95\% intervals over 160, 153 to 160 and 32 labels per step; the strip under (b) prints denominators below 160. Black segments: quarter means of step rates. Gray band: prompt-fix M-task replications, 32/128 and 35/128. Red ticks: retained positive-label records reporting geometric failure (19, 18 and 5; counts above one printed). Crosses: steps with missing retained files. Letters: recovered I/M task schedule. The rates summarize correlated logged labels within runs; independent task verification and seed replication remain future work (\cref{sec:boundary}).}
\label{fig:curves}
\end{figure}
\Cref{fig:curves} plots the logged positive-reward rate per training step for the three archived runs, binary M=5, dense M=5 and binary M=1, each a single trajectory of 20 steps with K=32 instructions per step. The task follows a recovered schedule shared by all three runs (two-cube I target at steps 1, 3, 4, 12, 13, 14, 16, 17 and 20; five-cube M target otherwise), so the first and last five-step windows have the same task composition, and the pooled binary rate is 537/1440 on I steps and 518/1760 on M steps. The binary run has 160 reward entries per step and rises from 45/160 at step 1 to 63/160 at step 20. The dense run's binary component is present for 3,170 of its 3,200 scheduled slots (153 to 160 per step); its first, last and peak step rates are 46/155, 60/158 and 63/159, and its 989 positive labels are 31.2\% of the observed slots and 30.9\% of the scheduled slots if every missing slot is counted as a failure. The gray band spans the prompt-fix baseline on the M task, 32/128 and 35/128 in its two replications (mean 26.2\%), run with an instruction-condensing call that the training runs do not have and at code-generation temperature 0.3 where the training runs used 0.1.

Averaged over the first and last five steps, the logged rate is 31.6\% and 35.3\% for the binary run (253/800 and 282/800), 30.0\% and 33.7\% for the dense run (pooled counts 238/794 and 268/795) and 25.0\% and 30.6\% for the M=1 run (40/160 and 49/160). These percentages are means of per-step rates; the dense pooled fractions differ slightly because its denominators vary. These increases describe logged training behavior within one trajectory per configuration, with shared instructions and tasks; independent held-out checkpoint evaluation remains future work. Wilson ribbons assume independent Bernoulli labels; estimating uncertainty across training seeds requires replicated runs.

The preliminary sweep's configurations differ from the central scheme in more than one factor at a time (its reference configuration uses K=128 and 10 steps where the central scheme uses K=32 and 20), and the single-epoch and E=3 runs also differ in execution fixes and provenance, so neither comparison isolates learning-rate or epoch effects.

A future matched design would fix M and the pipeline across three arms (low-LR/E=3, low-LR/E=1 and high-LR/E=3), plus high-LR/E=1 for the interaction. The archive lacks these matched arms. The M=5 and M=1 runs compare reward estimators at different downstream sampling costs; equal-compute comparisons remain future work.

A five-step M=1 record in the RQ2 archive is a byte-identical copy of the first five steps of the M=1 run and is counted once.

\section{Traceable Instructions and Failure Analysis}
\label{sec:examples}
Four instruction samples from step 20 of the main binary run show what the intermediate makes visible. The pick-and-place request in every pick-and-place run is: ``Fetch the red cubics from their original spots (on the table at the back) and then place them in the same disposal area (the central point of the table on the right).'' The stage-2 request asks for specific logical control instructions for the parameters and three core motion-planning functions. In the RL runs, stage 3 receives the base API and task code, the full sampled instruction, the completion request and the API addendum reproduced in Appendix~\ref{app:prompts}. Appendix~\ref{app:prompts} reproduces the task requests, the stage requests, the API addendum and excerpts of one traced step-20 record byte-for-byte from the archived files.

\begin{samepage}
Sample 000 of that step includes the following instruction excerpt:
\begin{Verbatim}[fontsize=\footnotesize,breaklines=true,breakanywhere=true]
B. fetching_cube (State Machine Steps 2, 3, 4)
- State 2 (Approach):
  - Define target_pos slightly above the cube (z + 0.02).
  - Generate trajectory to move arm to this position.
  - On completion (terminated), move to State 3.
- State 3 (Grasp):
  - Check closeGripper.
  - If successful, move to State 4.
\end{Verbatim}
\end{samepage}
The wording is retained with Markdown formatting and indentation normalized. Its five logged rewards are zero; retained code has a bracket-mismatch SyntaxError. This record separates the readable instruction from the syntax error in its generated program.

Three other samples from the same step expose distinct evidence types. Sample 001 has rewards (0, 1, 1, 0, 0), with a retained geometric-success message and maximum matched distance 0.0005\,m; independent task verification is pending. Sample 002 has five zeros and a retained NameError for undefined \texttt{pattern\_B}. Sample 026 has five ones but its retained log reports distance 0.9515\,m against 0.05\,m tolerance and failed placement. Because only the first reward-maximizing program was retained (\cref{sec:boundary}), the contradiction is established for that program; the other four programs behind the five rewards were not saved.

The 32 instruction outputs at this step contain 152--345 lines. With fence markers in the denominator but excluded from code interiors, median code-line fraction is 79.3\%; 26 first lines are distinct. All 32 code interiors and all 32 outside-code texts have different hashes. The 32 instructions therefore differ textually in both prose and code; behavioral diversity would need an execution-level comparison.

\section{Discussion, Limitations and Reproducibility}
The retained records support two diagnostic contributions. The intermediate instruction makes syntax errors, undefined names and discrepancies between rewards and execution logs traceable to specific instructions and programs. The reward audit identifies 19 positive-label records with reported geometric failure in the binary run and 18 in the dense run. These findings motivate independent task verification and distinguish changes in logged reward from improvements in validated task performance.

The study focuses on simulated manipulation. The RQ1 images serve as prompts rather than as a perception benchmark, and the RL runs are text-only. Hardware transfer would require camera and robot calibration, perception and state-estimation testing, contact and actuation robustness, latency measurement, and a separate safety review. Evaluation covers a subset of the 132 defined transformations; tests of the remaining tasks and generalization to held-out prompts, scenes and hardware remain future work.

The reproducibility scripts hash the evidence files they consume and regenerate the recounted values, traced examples and \cref{fig:curves} without model calls, and contain the CPU tests for geometric matching, tolerance boundaries, counts and incomplete statistical windows, together with the protocol checks of \cref{sec:boundary}. Further work includes integrating model-generated programs with the independent protocol, validating the target host, and establishing archived model and checkpoint provenance.

The pipeline's cost has five components: local instruction-policy training, hosted code generation, inference retries, simulator attempts and auxiliary parsing. Recovered code-generation ledgers total \$6.88 for binary M=5, \$6.95 for dense M=5 and \$1.40 for M=1 (prefix and resumed segment counted once); these are partial API charges and exclude local training compute, and the 160 A100-hour and 200 USD aggregates are estimates not reconciled with the recovered ledgers.

\section{Conclusion}
ICMC exposes the instruction between task parameters and frozen code synthesis, so execution outcomes can be traced to specific instructions and programs. The main training trajectory shows a rise in logged positive-reward rate from 31.6\% to 35.3\%. Auditing program-written verdicts also identifies at least 19, 18 and 5 positive rewards with retained logs reporting geometric failure across the three runs, establishing the need to distinguish logged rates from validated success. The independent protocol supports reproducible evaluation: its CPU implementation passed ten negative controls, and a hand-authored reference controller completed all 40 predeclared scenes. Next steps include evaluating model-generated programs under the independent protocol, replicating training across seeds, matching learning-rate/epoch comparisons and expanding o1 samples.

\section*{Acknowledgment and AI Assistance}
An AI assistant was used to debug and refactor experiment code for efficiency. The models studied (o1, o4-mini, Qwen3.5-35B-A3B and Qwen3-Coder-Next) are evaluated or trained as part of the experimental design; their generated instructions and programs are experimental outputs. The authors take responsibility for all content, evidence and citations.

\useRomanappendicesfalse
\appendices
\section{Recovered Prompt Components and a Traced Record}
\label{app:prompts}
Listings reproduce the cited archived line ranges without edits; original spelling is retained and long lines wrap.

\subsection{Task request strings}
Pick-and-place (\path{prompts/rq1-pick-place-user.txt}, line 3, also line 3 of both RQ2 I/M instruction templates):
\begin{Verbatim}[fontsize=\footnotesize,breaklines=true,breakanywhere=true]
[Task]: Fetch the red cubics from their original spots (on the table at the back) and then place them in the same disposal area (the central point of the table on the right).
\end{Verbatim}
Reshaping (\path{prompts/rq1-reshaping-user.txt}, line 3, also line 3 of both RQ2 reshaping templates):
\begin{Verbatim}[fontsize=\footnotesize,breaklines=true,breakanywhere=true]
[Task]: First fetch the red cubics from their original spots (on the table at the back) and then place them in the same disposal area (the central point of the table on the right) in the shape given in pattern_A. Then remove the dummy cubics and add new cubics to format the new shape given in pattern_B.
\end{Verbatim}

\subsection{Stage requests}
The 158-line RL stage-2 template \path{prompts/rq2-I-instruction-user.txt} refers to an image and arm-control task (line 1), gives the task string (3), and introduces the 140-line reference code (5, 7--146). The text-only runs supplied no image. Lines 148--158 contain the parameters and instruction request:
\begin{Verbatim}[fontsize=\footnotesize,breaklines=true,breakanywhere=true]
[Motion Control Logistic Paramters]
env = gymnasium.make('manipulator_mujoco/TestEnv-v0', render_mode='rgb_array')
observation, info = env.reset(seed=42)
robotic_arm = np.array([0.3, 0, 0.8, 0, 0, 0, 1]).reshape(1, -1)
cube_id = 0 
steps = 1 
middle_step = 1 
break_flag = 0 
terminated = env.step(robotic_arm)[2]

Please generate the specific [Logical Control Instructions] for each parameters in [Motion Control Logistic Paramters] regarding the whole [Basic operational codes], especially for the ## three core functions for motion planning.
\end{Verbatim}
The RL stage-3 builder (\path{prompts/rq2-code-builder-source.txt}, lines 23--30) uses the stage-2 template without line 158 as \path{trimmed_prompt} and the entire sampled instruction as \path{logical_suggestions}. The denoised driver appends the API addendum below and sends one user message.
\begin{Verbatim}[fontsize=\footnotesize,breaklines=true,breakanywhere=true]
    code_prompt = f"""{trimmed_prompt}

[Logical Control Instructions]:
{logical_suggestions}

Based on the above logical control instructions, please complete the Python code.
Return the complete code with all functions implemented.
Please do not change the functions in [Basic operational codes] unless you must do that."""
\end{Verbatim}
RQ1 uses a different request/separator (\path{prompts/rq1-code-builder-source.txt}, lines 11--13) and an image whose original bytes are absent from the retained archive:
\begin{Verbatim}[fontsize=\footnotesize,breaklines=true,breakanywhere=true]
    sub_task='''\nBased on the [Basic operational codes] and the [Logical Control Instructions], return the completed version of the [Basic operational codes].'''

    new_text = new_task_description + '\n\n[Logical Control Instructions]\n\n' + generated_logical_suggestion + sub_task
\end{Verbatim}

\subsection{API-documentation addendum}
\path{prompts/rq2-api-addendum.txt} stores the denoised driver's exact appended Python string; lines 1--2 are blank, and 3--27 follow:
\begin{Verbatim}[fontsize=\footnotesize,breaklines=true,breakanywhere=true]
[CRITICAL API NOTES — Read carefully before writing code]:

1. env.step() expects a 2D trajectory array of shape (N, 8) where column 0 is time
   and columns 1-7 are the pose [x, y, z, qx, qy, qz, qw].
   Do NOT iterate over trajectory rows. Pass the full trajectory at once:
       terminated = env.step(robotic_arm)[2]
   NOT: for row in robotic_arm: env.step(row[1:])  # WRONG — causes IndexError

2. The main control loop must call fetching_cube() and move_to_pattern() REPEATEDLY
   in a while loop. Each call advances middle_step by one phase when terminated==True.
   Do NOT call these functions just once — loop until cube_id >= pattern_B.shape[0].
   Example pattern:
       while cube_id < pattern_B.shape[0]:
           if steps == 1:
               pos, robot_pos, steps, cube_id, break_flag = starting_perception(cube_id, pattern_B, env, steps)
               if break_flag: break
           elif steps == 2:
               pos, robot_pos, steps, middle_step, robotic_arm = fetching_cube(pos, robot_pos, env, middle_step, robotic_arm, terminated, steps)
           elif steps == 3:
               robot_pos, steps, middle_step, robotic_arm, cube_id = move_to_pattern(cube_id, robot_pos, env, middle_step, robotic_arm, terminated, steps)
           terminated = env.step(robotic_arm)[2]

3. Use `pattern_B` consistently (the variable loaded from the .txt file).
   The get_target_pos() helper uses `pattern` — replace it with `pattern_B` or
   define `pattern = pattern_B` at the top.
\end{Verbatim}

\subsection{One traced record: step 20, instruction sample 001}
\path{examples/step020-rollout001} exactly copies the denoised-binary archive's \path{step_020/rollout_001}. Its \texttt{reward.json} records $[0,1,1,0,0]$ over $M=5$ programs (mean $0.4$); only sample 1, the first reward maximum, and its logs were retained. Lines 1--14 of the 229-line instruction follow:
\begin{Verbatim}[fontsize=\footnotesize,breaklines=true,breakanywhere=true]
Here are the specific **Logical Control Instructions** to complete the task of fetching red cubes from the back table and placing them in the central disposal area on the right.

These instructions fill in the missing logic within the `### starting the replacement` and `### ending the replacement` blocks and refine the control flow to ensure the `fetching_cube` and `move_to_pattern` functions are called sequentially for every cube.

### Logical Control Instructions

#### 1. Initialization & Pre-flight Check
Before entering the main loop, ensure the environment is stable and the initial state variables are correctly set to trigger the first perception phase.

*   **Action**: Reset the simulation time and verify the initial `robotic_arm` state.
*   **Logic**:
    *   Wait until `sim_time > 2.0` (as per your existing code) to ensure the arm has settled.
    *   Reset `cube_id` to `0`.
    *   Reset `steps` to `1` (indicating we are ready to start the perception phase).
\end{Verbatim}
In the 188-line \texttt{code.py}, lines 122--154 delimit the model-written main loop; 122--131 follow. The runner appends a placement checker (158--188) after \texttt{env.close()}. Absent from the raw model response, the checker shares the program's process.
\begin{Verbatim}[fontsize=\footnotesize,breaklines=true,breakanywhere=true]
### starting the replacement

# Main Loop Logic
while cube_id < pattern_B.shape[0]:
    # 1. Perception Phase: Identify cube location
    if steps == 1:
        pos, robot_pos, steps, cube_id, break_flag = starting_perception(cube_id, pattern_B, env, steps)
        # Safety check if cube_id exceeded
        if cube_id >= pattern_B.shape[0]:
            break
\end{Verbatim}
\path{exec_stderr.txt} is empty; the checker's six-line \path{exec_stdout.txt} follows:
\begin{Verbatim}[fontsize=\footnotesize,breaklines=true,breakanywhere=true]
planned_positions: [[ 0.4 -0.1]
 [ 0.6 -0.1]]
cube_positions: [[ 0.40042537 -0.09990171]
 [ 0.60047451 -0.09992712]]
Success: All cubes correctly placed! max_match_dist=0.0005
Pick-and-place routine completed successfully!
\end{Verbatim}
The success line and 0.0005\,m distance are this sample's only retained placement evidence; the other four programs/logs are unsaved. Independent checking would determine whether this logged positive meets task-success criteria (\cref{sec:boundary}). \Cref{sec:examples} contrasts step-20 samples 000, 002 and 026.

\subsection{Provenance boundary}
These recovered templates, builder source and outputs are user-text components, not complete API transcripts. The original trainer, chat serializer, any system message, tokenizer version and RQ1 image bytes are not available in the retained archive. Complete templates, four full step-20 records and a SHA-256 evidence manifest are retained for release.

\bibliographystyle{IEEEtran}
\bibliography{custom}
\end{document}